\documentclass{article}

\usepackage{PRIMEarxiv}

\usepackage[utf8]{inputenc} 
\usepackage[T1]{fontenc}    
\usepackage{hyperref}       
\usepackage{url}            
\usepackage{booktabs}       
\usepackage{amsthm, mathrsfs, amsmath,amssymb,amsfonts}       
\usepackage{nicefrac}       
\usepackage{microtype}      
\usepackage{lipsum}
\usepackage{fancyhdr}       
\usepackage{graphicx}       
\usepackage{xcolor}
\usepackage{algorithm}%
\usepackage{algorithmicx}%
\usepackage{algpseudocode}%
\usepackage{listings}%
\usepackage{manyfoot}%
\usepackage{multirow}%
\graphicspath{{media/}}     
\usepackage{array}

\usepackage{subcaption}
\usepackage{tabularx}
\usepackage{booktabs}
\usepackage{url}
\usepackage{enumitem}

\title{A New Hybrid Intelligent Approach for Multimodal Detection of Suspected Disinformation on TikTok}

\author{
  Jared D.T. Guerrero-Sosa \\
  Department of Technologies and Information Systems \\
  University of Castilla-La Mancha, Spain\\
  \texttt{JaredDavid.Guerrero@uclm.es} \\
  \And
  Andres Montoro-Montarroso \\
  Department of Technologies and Information Systems \\
  University of Castilla-La Mancha, Spain \\
  \texttt{Andres.Montoro@uclm.es} \\
  \And
  Francisco P. Romero \\
  Department of Technologies and Information Systems \\
  University of Castilla-La Mancha, Spain \\
  \texttt{FranciscoP.Romero@uclm.es} \\
  \And
  Jesus Serrano-Guerrero\\
  Department of Technologies and Information Systems \\
  University of Castilla-La Mancha, Spain \\
  \texttt{Jesus.Serrano@uclm.es} \\
  \And
  Jose A. Olivas \\
  Department of Technologies and Information Systems \\
  University of Castilla-La Mancha, Spain \\
  \texttt{JoseAngel.Olivas@uclm.es} \\
}

\begin{document}
\maketitle

\begin{abstract}
In the context of the rapid dissemination of multimedia content, identifying disinformation on social media platforms such as TikTok represents a significant challenge. This study introduces a hybrid framework that combines the computational power of deep learning with the interpretability of fuzzy logic to detect suspected disinformation in TikTok videos. The methodology is comprised of two core components: a multimodal feature analyser that extracts and evaluates data from text, audio, and video; and a multimodal disinformation detector based on fuzzy logic. These systems operate in conjunction to evaluate the suspicion of spreading disinformation, drawing on human behavioural cues such as body language, speech patterns, and text coherence. Two experiments were conducted: one focusing on context-specific disinformation and the other on the scalability of the model across broader topics. For each video evaluated,  high-quality, comprehensive, well-structured reports are generated, providing a detailed view of the disinformation behaviours.
\end{abstract}


\section{Introduction}
The spread of disinformation on multimedia platforms is a growing problem with significant political, health, and social consequences \cite{Muhammed2022}. The sophisticated use of text, images, and videos often creates convincing narratives that disrupt the truth and manipulate public opinion. Such misleading content, amplified by the algorithms of social media platforms like YouTube, Facebook, Instagram, and TikTok, spreads quickly and pervasively, instantly reaching global audiences. This uncontrolled spread of disinformation poses significant threats, including political manipulation, incitement of violence, public health crises, and undermining democratic processes. The challenge lies in developing effective strategies and tools to identify and counteract these disinformation campaigns without infringing upon free speech and individual privacy rights \cite{delVicario2016}.

Identifying disinformation behaviours in video content is multifaceted, employing techniques that analyse both nonverbal cues and explicit verbal content. Standard methods include observing body language and facial expressions, tracking movement patterns, and monitoring behavioural changes. Specific facial actions associated with genuine expressions can serve as valuable indicators, as they are challenging to inhibit or fake~\cite{Su2016}. Machine learning (ML) techniques are increasingly used to detect deceptive behaviours. However, their performance can be affected in the context of social networks, as they may inadvertently perpetuate biases in training datasets.

In recent years, analysing multimedia content involving human behaviour has become an extraordinary phenomenon~\cite{Nguyen2022}. For example, Choi and Ko~\cite{choi2022} presents an approach to detect disinformation videos using domain knowledge, considering the potential meaning of comments to detect misleading videos, and multimodal data fusion through a linear combination to adjust the encoding rate for each video characteristic.  Alternatively, the research by Bonomi et al.~\cite{Bonomi2021}  suggests a significant correlation between social network users who disseminate false information and adverse traits such as narcissism, Machiavellianism, and psychopathy. Consequently, identifying fraudulent content on social networks has become a vital research domain. The main focus of this area encompasses psychological factors, algorithmic techniques, and representative datasets, underscoring its interdisciplinary nature~\cite{Shu2017}.

This work aims to propose and develop a novel hybrid intelligence framework that unifies the computational power of deep learning with the interpretative capabilities of fuzzy logic techniques. The main goal of the proposed framework aims to facilitate the objective detection of human behaviour in multimodal content, specifically the identification of multimodal indications of disinformation. This system comprises two core elements: a multimodal feature analyser and a multimodal disinformation content detector. The feature analyser leverages and employs advanced Deep Learning techniques to extract multimodal - text, image, and audio - features from multimedia pieces, providing measures related to human traits for detecting disinformation. The system then employs a Fuzzy Logic-based detector to interpret these measures, identifying recurring behaviours and patterns common among disinformation spreaders, thereby enhancing the accuracy of disinformation detection. Following the preceding findings \cite{Guerrero-Sosa2023}, the fusion of Deep Learning and Fuzzy Logic facilitates a rich, multimodal analysis, offering a comprehensive understanding of the analysed content. Notably, the framework also ensures explainability, providing transparent and comprehensive insights into the detection process and the basis for the system's determinations. It is our contention that this represents the inaugural effort to employ a multimodal methodology for the purpose of identifying purported disinformation in TikTok videos with an emphasis on explainability.

The main contributions of this article are as follows:
\begin{itemize}
\item A multimodal detection of disinformation on TikTok based on Granular Linguistic Model of Phenomena and Psychosociological Traits is proposed, with a particular focus on the Big-5 behavioural framework.
\item A Hierarchical Perception Mapping is defined to capture and model more complex phenomena for interpretability and flexibility.
\item A Prompt Generator is designed for capturing the essence of the perceptions at different levels of granularity.
\end{itemize}

The organisation of this paper is detailed below: 
Section \ref{sec:related_work} broadly looks at recent progress in studying multimodal disinformation analysis, forming the backdrop for this research. Next, in Section \ref{sec:gnlp} explains the main theories that support the suggestion, including fuzzy elements and the Computer Theory of Perceptions. Section \ref{sec:meth} outlines the two-step approach to multimodal analysis used in this study. Section \ref{sec:ilustrative_example} illustrates the various perceptions and levels of granularity of behaviour related to the suspicion of disinformation to enhance the methodological approach's intelligibility. 
Section \ref{sec:experiments} presents an in-depth analysis of two experiments involving TikTok videos. The objective is twofold: firstly, to examine whether disinformation users can be identified in specific contexts; secondly, to ascertain whether the method can be scaled to address a widespread topic with a large volume of potential disinformation videos. Finally, section \ref{sec:concl} consists of findings and points out areas that could benefit from more research in the future.

\section{Related Work}
\label{sec:related_work}

The field of multimodal fake news detection has been significantly advanced through the integration of a variety of modalities, including text, images, and videos. This integration has expanded the scope of detection methods beyond those based entirely on text \cite{Comito2023}. A review of the literature on deep learning techniques reveals that while considerable progress has been made in utilising multimodal data, challenges remain in the effective fusion of data and the integration of underutilised modalities, such as news propagation networks \cite{Tufchi2023}. Furthermore, research gaps remain in the real-time detection and evaluation of metrics, with further opportunities to enhance model explainability and incorporate more sophisticated machine-learning approaches. Additionally, the interdisciplinary nature of disinformation research, involving computer science and the humanities, underscores the need for unified terminology and a more cohesive research community to address issues such as temporal dynamics in the spread of disinformation \cite{Wilson2023}.

Disinformation detection based on textual content analysis involves examining linguistic attributes, including lexical, syntactic, and semantic features, using Natural Language Processing (NLP) techniques. The primary goal is to identify unique stylistic patterns used by disinformers \cite{MontoroMontarroso2023}, based on the idea that these entities typically adopt specific writing styles to attract and convince social media users, thereby instilling a false sense of trust. However, disinformers can mimic the writing styles of credible sources. Schuster et al. \cite{Schuster2020} found that while models based on stylistic features effectively detect human-written disinformation, they struggle with synthetic texts generated by advanced language models. Research on disinformation in social networks covers various domains. For instance, Peng et al. \cite{PENG2023120501} investigated the dissemination of false information about COVID-19, using a question-based learning approach to evaluate the reliability of online texts. In a separate study, Rastogi and Bansal \cite{Rastogi2022} proposed an integrated methodology for the detection of disinformation on social media. This approach combines style-based and social context-based features and demonstrates the efficacy of an ensemble model in differentiating between authentic news, disinformation, and satire.

Due to the multimedia capabilities of social media platforms, malicious users create disinformation using text, images, and videos to enhance the content's credibility. Recently, these multimedia elements have been utilised to improve the accuracy of ML models \cite{Hangloo2022}. This method, known as multimodal disinformation analysis, involves various techniques to address the problem. For instance, Singh et al. \cite{singh2021} proposed a multimodal approach that integrates textual and visual features from labelled news article datasets to detect disinformation automatically, demonstrating that combining classic ML algorithms and multimodal data yields better results than relying on a single data type.
 c
Deep Learning techniques have become increasingly popular in multimodal disinformation detection because they effectively handle sequential data (text) and structured data with known topology (images). For instance, Chai et al. \cite{CHAI2024121588} introduced a model that combines traditional machine learning interpretability with the representational power of Deep Learning to identify misleading reviews and fraudulent emails. Typically, Deep Learning-based models for multimodal disinformation detection use two neural network architectures: text and images. These architectures extract features combined using a final neural network \cite{Sengan2023, Jing2023, Ghorbanpour2023}. Recent advancements have incorporated the attention mechanism \cite{Guo2023}, which allows the model to focus on the most relevant aspects of the input data, enhancing detection accuracy \cite{Yadav2023}. Additionally, Hua et al. \cite{Hua2023} incorporated a contrastive learning module alongside attention mechanisms, which learns data representation by contrasting example pairs. This approach compares similar past news articles to improve the performance of the disinformation detection model.

The feature fusion not only occurs at the end of the multimodal disinformation detection process. Meel and Vishwakarma \cite{Meel2023} distinguished tree approaches to fusion: early fusion ---feature vectors from each pipeline are merged by concatenating them together \cite{Xiong2023}--- late fusion ---involves combining the final probability decisions from each pipeline \cite{Singh2023}--- and hybrid multi-level fusion  --- encompasses both early and late fusion, distributing the inputs among these two approaches. 

Other works try to enhance the detection of disinformation by incorporating additional features. For example, Wu et al. \cite{Wu2023} introduced the Human Cognition-based Consistency Inference Networks (HCCIN) model, which captures consistent and inconsistent information between news content and its comments, incorporating an extra context element. On the other hand, Giachanou et al. \cite{Giachanou2020} and Li et al. \cite{Li2022} presented a multimodal system grounded on a neural network that integrates textual, visual, and semantic data to distinguish between false and trustworthy information. Furthermore, Zhang et al. \cite{Zhang2022} focused on extracting semantic information from images by analysing visual scenes such as location, seasons, and weather. Similarly, Wang et al. \cite{Wang2023} incorporated entities extracted from images as additional features in a multimodal approach.

A different approach was proposed by Gôlo et al. ~\cite{Silva2023}. In the work, a novel method that utilises Multimodal Variational Autoencoders (MVAEs) to learn a new representation by combining various modalities such as text embedding, topic, and linguistic information is described. Notably, their employed learning method is One-Class Learning (OCL). This ML approach only requires positive examples to build a classifier, eliminating counterexamples and reducing the effort required for data labelling.

According to Hameleers et al. \cite{Hameleers2020}, multimodal disinformation is perceived as slightly more credible than textual disinformation. Moreover, to enhance the robustness of disinformation detectors, Chen et al. \cite{Chen2023} recommended giving greater emphasis to visual features, particularly images associated with trending events, as these visual elements represent a shared vulnerability among such detectors.

Conversely, Han et al. \cite{Han2024} presents the Multifaceted Reasoning Network for Explainable Fake News Detection (MRE-FND), which integrates a multitude of data sources, employs a dual graph neural network, and utilises explainable reasoning to enhance interpretability and accuracy. This model demonstrates the capacity for noise elimination and attains a state-of-the-art performance on major datasets. In contrast, Peng et al. \cite{Peng2024} proposes Contextual Semantic Representation Learning for Multimodal Fake News Detection (CSFND), which emphasises the integration of contextual semantics into multimodal data, addressing inconsistencies in detection by incorporating an unsupervised context learning stage and a contextual testing strategy, enhancing detection accuracy and robustness.

In the work of Huda et al. \cite{Huda2024}, Fake-checker, a method that fuses deep learning with texture features, using the novel DMLHP descriptor and Inception V3 for feature extraction, is introduced. This approach achieves high accuracy in detecting deep fakes and targeting face-swapping and face-re-enactment techniques to combat disinformation.

\section{Granular Linguistic Model of Phenomena}
\label{sec:gnlp}

A multi-level approach is frequently employed to describe complex behaviours effectively. The concept of granularity, which groups data into units of information called granules, is paramount in capturing the nuances of a phenomenon in its specific context \cite{zadeh2002granular}.

An extension to this model is proposed by introducing Hierarchical Perception Mapping (HPM), which introduces a hierarchical and recursive structure to Perception Mapping (PM). This enhancement permits the capture and modelling of more complex phenomena, thereby facilitating greater granularity and flexibility in the representation of perceptions.

In essence, the Computational Perceptions (CPs) provide a detailed analysis of specific aspects of the system under examination. In modelling disinformation behaviour in a video, several aspects are examined based on subjective perceptions.

\begin{figure}[htp!]
\centering
\includegraphics[scale=0.4]{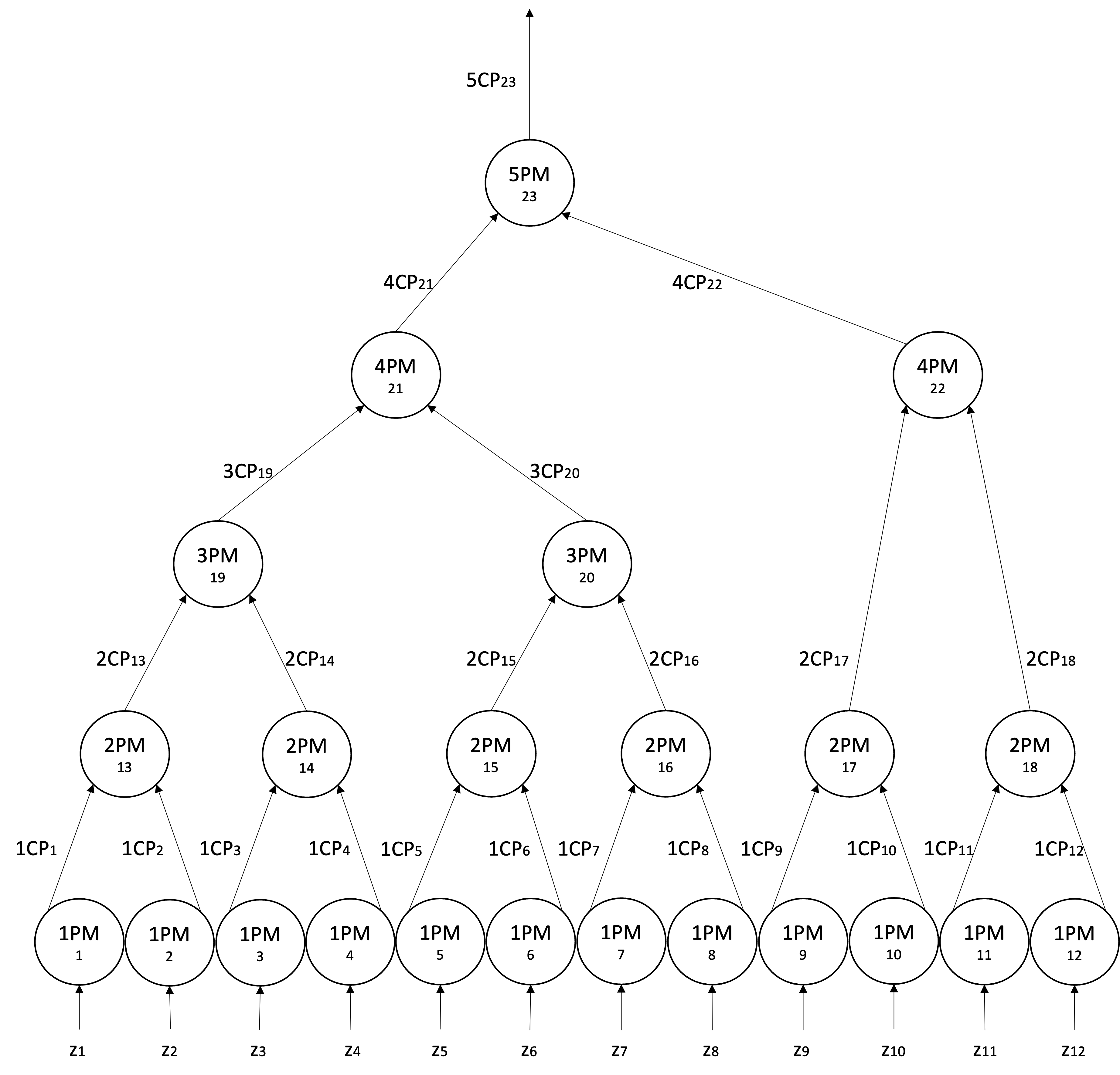}
\caption{Example of a simple GLMP.} \label{fig:gmlp}
\end{figure}

A Granular Linguistic Model of Phenomena (GLMP) displays information in a granular way across a network of PMs, contingent on the requisite level of detail. A CP at one level may serve as the input for a subsequent level, with the potential for multiple levels of recursion. The PM network maps input and output nodes and CPs through edges. At higher levels of analysis, the aggregation of CPs by PMs facilitates the generation of new CPs, thereby enhancing the representation and comprehension of the phenomenon. An example is shown in Fig.~\ref{fig:gmlp}. In the following, the components of PMs and CPs will be described based on the definitions initiated by Trivino and Sugeno \cite{TRIVINO201322} and updated by Conde-Clemente et al. \cite{CONDECLEMENTE201746} and de Anda-Trasviña et al. \cite{de2022natural}.

\newpage

A Computational Perception ($CP$) is represented as a tuple $(A, W, R)$, where:

\begin{itemize}
\item $A$: $(a_1, a_2, \dots, a_n)$ is a vector of linguistic expressions, such as words or sentences in natural language, encompassing the entire linguistic domain of the $CP$. Each component $a_i$ corresponds to the most appropriate linguistic value of the $CP$ for a specific behaviour, with a particular degree of granularity. For instance, the perception level of extroversion in the person who sent the message could be modelled using $A = (low, medium, high)$.

\item $W$: $(w_1, w_2, \dots, w_n)$ is a vector with validity degrees $w_i \in [0, 1]$. Each validity value $w_i$ represents the degree to which the linguistic expression $a_i$ accurately describes the specific input data. The sum of all validity degrees must be $\sum{w_i} = 1$.

\item $R$: $(r_1, r_2, \dots, r_n)$ represents a relevance vector, where each $r_i \in [0, 1]$ signifies the relevance degree associated with the linguistic expression $a_i$ within the specified context. The designer assigns the value of each $r_i$, and it's noteworthy that the values of CP may vary based on the distinct types of users.

\end{itemize}

Perception Mappings ($PM$) are essential in creating and aggregating $CPs$. Each $PM$ combines a set of input $CPs$ to form a unified $CP$. A $PM$ is represented as a tuple $(U, y, g, T)$, where:

\begin{itemize}
\item $U$: $ (u_1, u_2, \dots, u_n)$ is a vector comprising $n$ input $CPs$, denoted as $u_i = (A_{ui}, W_{ui}, R_{ui})$. In the case of first-level Perception Mappings ($1PM$), the inputs can be numerical values ($z$ $\in$ $R$) obtained from measurement procedures.

\item $y$: $(A_y, W_y, R_y)$ represents the output $CP$ generated by the $PM$.

\item $g$ denotes the aggregation function used in the $PM$. In Fuzzy Logic, various aggregation functions have been developed to handle different linguistic expressions effectively. For $1PM$, $g$ is constructed by utilising a set of membership functions. An aggregation function can calculate the degree of validity ($g_W$) or relevance ($g_R$).

\item $T$ represents a text generation algorithm that enables the generation of all possible sentences associated with the linguistic expressions in $A_y$. $T$ is a linguistic template encompassing a set of potential linguistic expressions.

\end{itemize}

In Hierarchical Perception Mapping (HPM), a map is presented as a tuple $(F, E, V, D)$, where:

\begin{itemize}
\item $F$: A linguistic label represents the specific phenomenon or sub-phenomenon being modelled. It is a key descriptor that provides a meaningful name for the overall perception captured by the HPM.
\item $E$: Represents the event of data being processed or evaluated. This can range from raw input data in the case of first-order HPMs to aggregated data from lower-order HPMs.
\item $V$: Represents the value associated with the HPM, typically obtained through evaluation or processing of the event or data. This value essentially quantifies the perception related to $F$.
\item $D$: Represents the set of dependent or lower-order HPMs. Each element of $D$ is a tuple in the form of an HPM, thus providing a hierarchical and recursive structure. It allows us to capture and model perceptions composed of sub-perceptions or sub-phenomena.
\end{itemize}

The Prompt Generator (PG) function is introduced as an essential component that facilitates the generation of effective prompts for text generation. The PG function $PG: HPM \rightarrow Prompts$ takes an HPM as input and produces prompts that guide the text generation process. By leveraging the hierarchical and recursive structure of the HPM, the PG function can extract meaningful information and generate prompts that capture the essence of the perceptions at different levels of granularity. These prompts act as valuable cues for the subsequent text-generation step.

\section{Methodological Proposal}
\label{sec:meth}

\begin{figure}[ht!]
\centering
\includegraphics[scale=0.25]{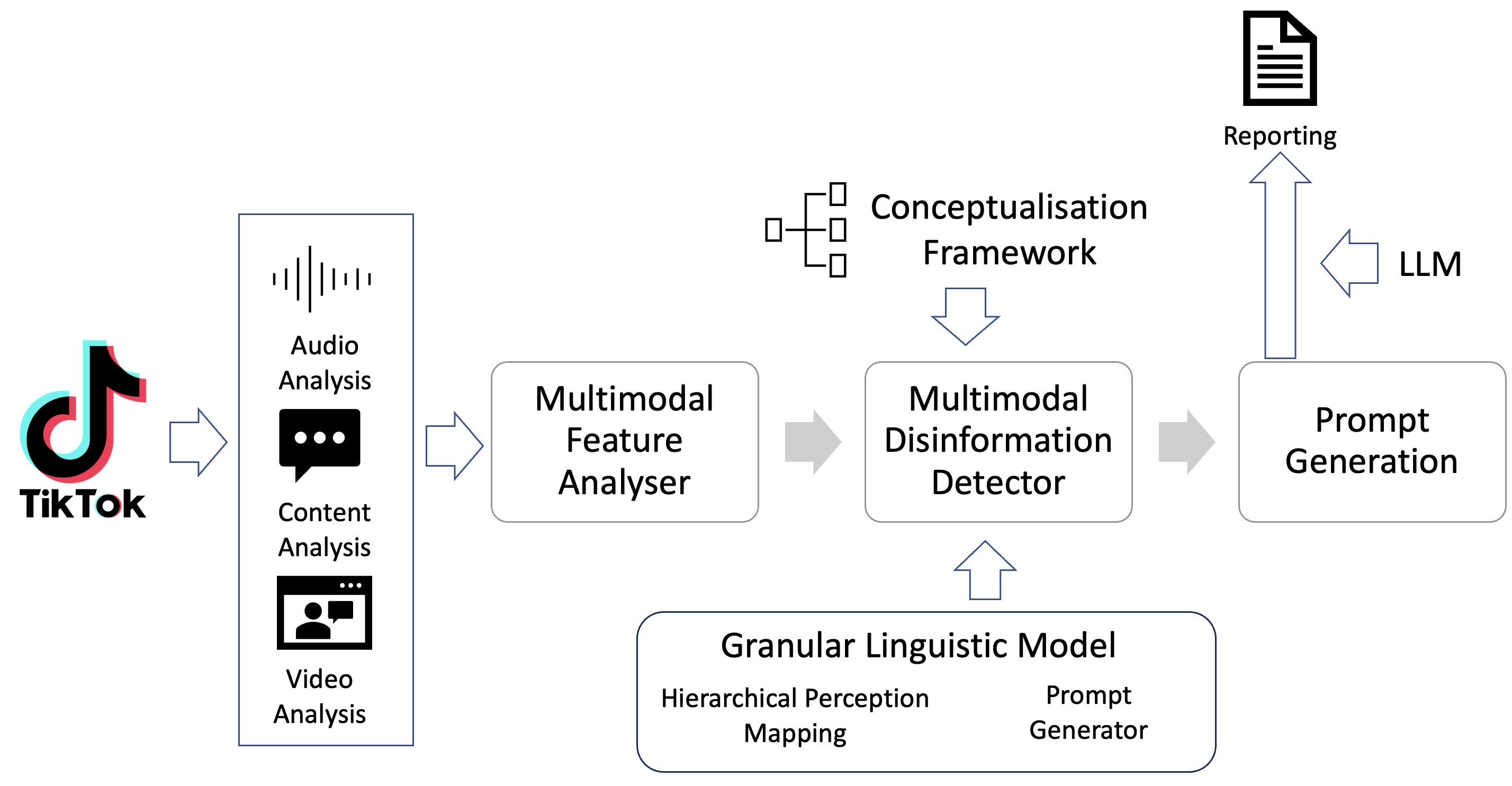}
\caption{Overview of the proposed approach.}
\label{fig:overview}
\end{figure}

The process is divided into four components (Fig.~\ref{fig:overview}) described in the following subsections.

\subsection{Conceptualisation Framework}

A conceptualisation framework is a methodological approach designed to study and analyse multimedia data to identify the most relevant characteristics of behaviour to be detected in video footage. The framework is constructed through collaborative work, where a panel of experts provides the dimensions and attributes of each behaviour, and an engineering team defines the tools to measure each attribute and the aggregation procedures. The framework aims to establish a structured and systematic process for identifying and characterising behaviours of interest in multimedia data, which can then be used to develop more accurate and reliable automated systems for behaviour detection and analysis. By providing a standardised and consistent methodology for conceptualising behaviours, the framework helps ensure that all relevant dimensions and attributes are considered and that the resulting system is valid and effective.

Each concept represents a $CP$, and then a membership function and an aggregation process, such as a set of rules, are associated with it. The aggregation process is evaluated, and conclusions are obtained as $PM$s. Measures are $1PM$, and attributes and dimensions are aggregations of $CP$s.  

\begin{figure}[!htp]
\centering
\includegraphics[scale=0.08]{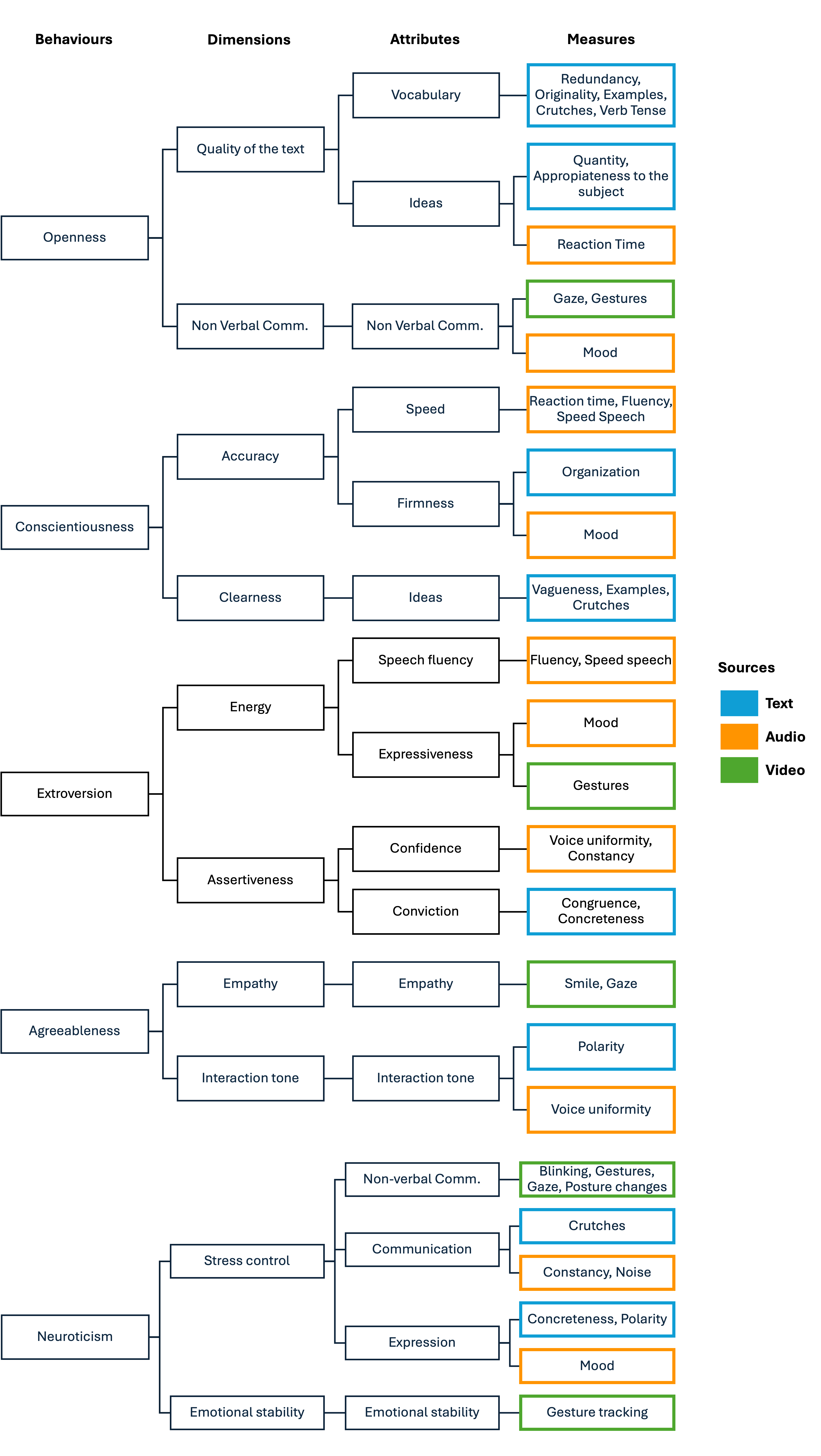}
\caption{Dimensions, attributes and measures to detect suspected disinformation.}
\label{fig:tab_concept}
\end{figure}

The video footage analysis identifies psychosociological traits that can accurately predict suspicion of spreading disinformation. As described later in \ref{subsec:multimodal_disinformation_interpreter}, the classification known as \textit{Big-5} was used, which describes individual differences in behaviour and thinking~\cite{novikova2019five}. Then, the conceptualisation framework of this specific behaviour is divided into different dimensions/attributes (provided by a panel of experts) and measures (see Fig.~\ref{fig:tab_concept}). 

The suggested conceptual framework draws from Knowledge Engineering and Machine Learning in a combined approach to examining and evaluating multimedia data. It aims to identify the essential features of the behaviours under investigation. Subsequently, the engineering team determines the appropriate tools for quantifying each characteristic.

Using the GLMP, the metrics' values can be represented with linguistic labels to obtain the defined attributes, dimensions, and behaviours corresponding to the Perceptual Mappings and Conceptual Perceptions.

\subsection{Multimodal Feature Analyser}

This component assumes responsibility for computing the first-order perceptions ($1PM$) involved in the procedure. It comprises a multitude of components based on ML techniques. Some of these elements consist of approaches such as speech transcription from videos, audio retrieval, or video segmentation into frames. Others constitute sophisticated tools founded on Convolutional Neural Networks capable of detecting emotions from audio or identifying blink/gaze patterns, as well as transformers and autoencoders to extract textual topics.

The information from the video content will undergo a cleaning and normalisation process. The video and audio will be divided into more manageable sections and then transformed into MP4 and WAV file formats using the FFmpeg software. The following steps involve an individual analysis of the video and audio content.

\textbf{Audio Transcription:} The process leverages Whisper \cite{Radford2022b}, an advanced speech recognition model, to transcribe audio from disparate languages into a common language, such as English. Whisper also handles the segmentation of the audio into discrete units. Segmentation into segments facilitates comprehensive processing and extracting pertinent information, such as semantic and temporal content, for subsequent analysis.

\textbf{Audio Analysis:} The analysis of extracted audio is concerned with evaluating the acoustic properties of speech that may indicate disinformation through MyProsody. These properties include overlapping speech and high levels of entropy. By studying these acoustic patterns, speech characteristics that differ from native patterns can be identified \cite{Chen2018}. This provides clues about the authenticity and coherence of the content, thus enabling a better understanding of the nuances of the speech and the detection of possible manipulations or anomalies. The audio is analysed to obtain the values of the measures \textit{reaction time} (duration in seconds it takes to start the speech), \textit{mood} (reading, normal or passionate), \textit{fluency} (number of pauses and interruptions) and measures of the fundamental frequency of the voice (voice uniformity). 

\textbf{Text Analysis:} This task enables the interpretation of the transcribed content by identifying parts of speech and discourse markers, breaking down the text into grammatical and structural components. The semantic representation of the text makes it possible to understand the context and meaning, which facilitates the identification of patterns related to the spread of disinformation. The Spacy library \footnote{\url{https://spacy.io}} was used for natural language processing (NLP). Analysis of the transcribed audio provides measures such as concreteness, organisation, crutches, vagueness and order of speech structure. In addition, originality - measured by the use of specific and unusual vocabulary - and redundancy in discourse - measured by the repetition of common words - were assessed using specific dictionaries. Part of Speech (POS) tagging determined the occurrence of adjectives and verb tense, while semantic analysis benefited from word representation techniques such as Word2Vec \cite{mikolov2017advances} and Global Vectors for Word Representation (GloVe) \cite{Pennington2014}. Finally, sentiment analysis techniques measured the polarity of speech by classifying it as \textit{positive}, \textit{negative}, or \textit{neutral}.

\textbf{Text Classification:} The classification procedure proves beneficial for quantifying the pertinence of the speech to the phenomenon to be studied. This is done by assessing the text's depth, coherence, and main elements associated with the theme. This method has been utilised in prior studies involving text analysis to identify specific behavioural patterns~\cite{Martinez-Castano2020}. Text's thematic categories were derived based on a hierarchical taxonomy for topic extraction and other metric calculations \cite{Ermakova2023}. The taxonomy classification depended on the text content (EuroVoc classification for Spanish texts, IPTC classification for other languages).

\textbf{Video Analysis:} The main goal is assessing visual elements that may indicate disinformation. According to Giannakakis et al. \cite{GIANNAKAKIS201789}, involuntary and semi-voluntary facial indicators are used for objective emotional evaluations. These elements include facial and body behaviour. Identifying signs of deception or manipulation, which complements the analysis of verbal and textual content, provides a more complete picture of the phenomenon of disinformation. OpenCV, a set of software resources designed for assessing and manipulating images and videos \footnote{\url{https://opencv.org}}, was employed to analyse the videos, detecting and tracking individuals and facial attributes such as eyes, lips, and faces. Moreover, the 68-point facial landmark detector \cite{6909637}, was utilised for identifying landmarks characterising the human face including measures of eye activities (gaze and blink), mouth movements, and head motion parameters, which serve as viable stress and anxiety markers. Gestures were detected using the DeepFace library \cite{6909616}, classifying expressions into \textit{angry}, \textit{sad}, \textit{neutral}, \textit{disgust}, \textit{surprise}, \textit{fear}, and \textit{happiness}.

\subsection{Multimodal Disinformation Detector}
\label{subsec:multimodal_disinformation_interpreter}

The proposed GLMP aims to produce linguistic reports of the analysed behaviour. Each level validates and generates sentences that describe the current condition according to a specific level of granularity. The first part indicates the status based on the results of the multimodal feature analyser.  Then, these values are aggregated using fuzzy procedures to represent different attributes or dimensions of the behaviour, and the last part analyses the level of suspicion of disinformation of the video.  

From second-order PMs to $n$-1-order PMs (where $n$ is the number of orders of the phenomenon to be represented), the aggregation function depends on the number of entries the PM will receive. The aggregation operator's heuristic depends on the characteristics of the input variables. If the complexity of the fuzzy rule set is acceptable, the analysis is performed using a fuzzy inference process; for more complex scenarios where fuzzy rules are not applicable, weighted average operators are used.

In particular, for the attributes \textit{vocabulary}, \textit{ideas}, \textit{non-verbal communication} (for both \textit{openness} and \textit{neuroticism}) and \textit{expression}, weighted average operators were employed, with the weight of each entry assigned by the expert panel and dependent on its relevance in representing an attribute.

For the remaining attributes, dimensions and behaviours, a set of fuzzy rules was defined, establishing the composition of each item at varying levels, primarily \textit{low}, \textit{medium} and \textit{high}.

Once the fuzzy values corresponding to each of the behaviours have been calculated using fuzzy aggregation techniques, the values are combined to obtain a single output indicating the suspicion of disinformation in a video, which can be interpreted more intuitively than the individual values of the behaviours. However, each behaviour's influence on disinformation differs depending on the domain to which the message belongs. 

Some psychological models broadly describe human behaviour characteristics, with the Big-5 being one of the most extended. This model identifies the five elementary characteristics of human beings: openness, conscientiousness, extroversion, agreeableness and neuroticism, and they are present to a greater or lesser degree, depending on the person \cite{Goldberg1992}. 

\begin{figure}[ht!]
\centering
\includegraphics[scale=0.25]{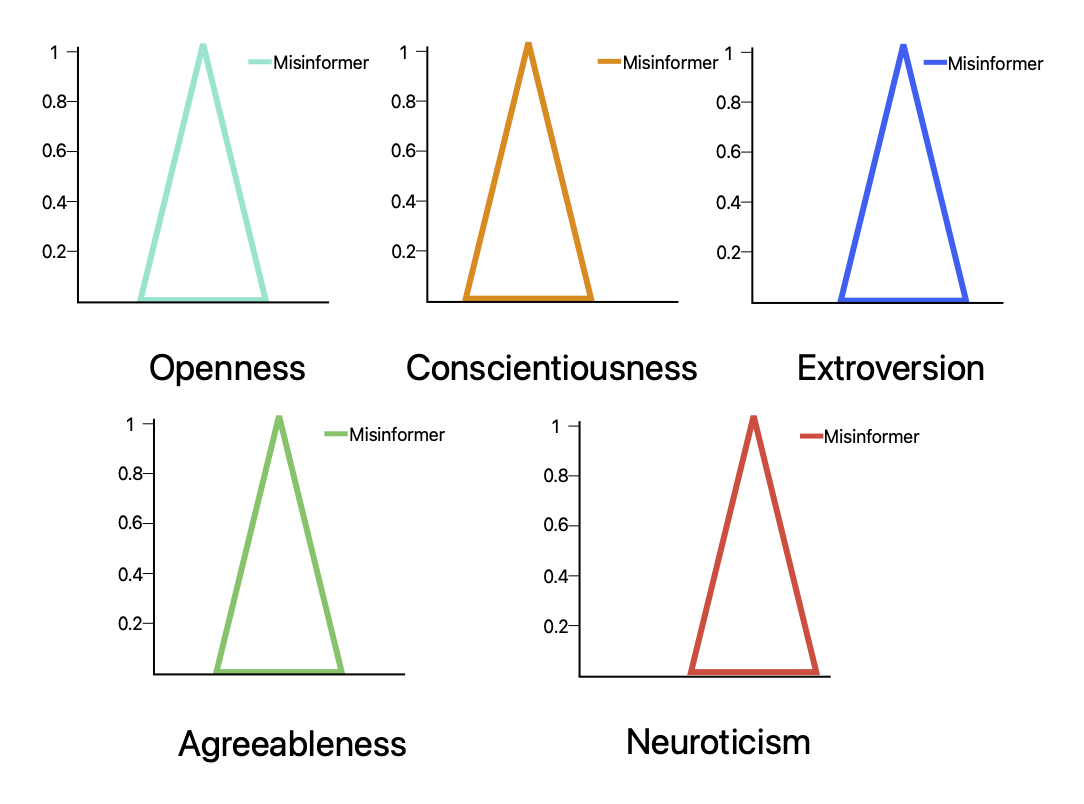}
\caption{Fuzzy sets of disinformer behaviours in the political domain.}
\label{fig:beahaviors}
\end{figure}

As the case study presented in Section \ref{sec:experiments} focuses on a dataset of TikTok videos related to the invasion of Ukraine and is about a political event, it is required to know the behaviours shared by people who spread disinformation in this domain, and from what was mentioned in the work of Srinivas et al. \cite{PYKL2022117952}, the spreaders of disinformation in the political domain are neurotic by nature based on the Big-5 model. Therefore, the scores mentioned by the author were used to define fuzzy sets (Fig. \ref{fig:beahaviors}) for each behaviour studied and detected the degree of belonging with the disinformer profile. 

This step is necessary, as it indicates the degree of membership of each person's behaviour during the video and how it is associated with the characteristics of a disinformer. Then, these values are combined to obtain a single output indicating the suspicion of disinformation in a video in the political domain using the Choquet Integral (Eq. \ref{choquet_integral}).

The steps to obtain the output using this aggregation function are:
\begin{itemize}
    \item Define all subsets from 1 to 4 elements and the complete set of evaluated behaviours.
    \item Assign the weight or importance of each subset to assess the suspicion of being a disinformer. The weight is between 0 and 1, and 1 is assigned only to the complete set. The experts assigned the weights according to the relevance of the combination of each subset of elements in the assessment.
    \item Sort the criteria obtained in ascending order and calculate the output.
\end{itemize} 

In this case, the behaviours are not independent but are interrelated and can influence each other. Once the criteria and the general fuzzy measure have been defined, the next step is to aggregate using the Choquet Integral, whose advantage as an aggregation function is the usefulness of combining groups of inputs rather than considering the importance or magnitude of individual inputs \cite{Beliakov2016}. The Choquet Integral formula is presented in Eq. \ref{choquet_integral}.
    
    \begin{equation}
        C_v(x) = \sum_{i=1}^{n} [x_i - x_{i-1}]v(H_i)
        \label{choquet_integral}
    \end{equation}
    where:
    \begin{itemize}
        \item $\left\lbrace x_1, x_2, x_3,...,x_n\right\rbrace$ is a set of criteria ordered in ascending order
        \item $x_0=0$
        \item $v$ is a fuzzy measure
        \item $H_i=\left\lbrace i,...,n \right\rbrace$ is the subset of indices of the $n-i+1$ largest components of $x$
    \end{itemize}

\begin{figure}[ht!]
\centering
\includegraphics[scale=0.4]{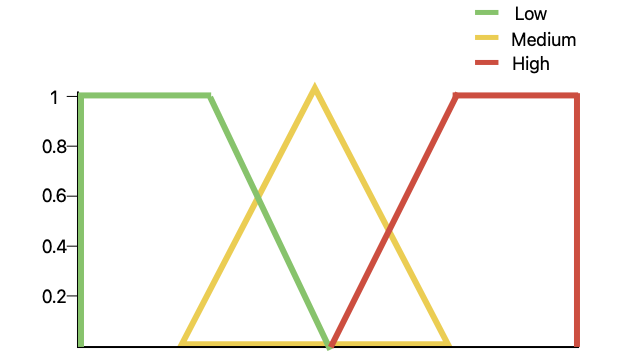}
\caption{Fuzzy sets of suspicion of disinformation.}
\label{fig:disinformation_fuzzy_sets}
\end{figure}

$C_v(x)$ is a value in the [0, 1] interval that indicates that the higher it is, the higher the suspicion of disinformation in the video evaluated. For this, three fuzzy sets have been defined to linguistically represent the degree of suspicion of disinformation (Fig. \ref{fig:disinformation_fuzzy_sets}).

\subsection{Prompt Generation}

This module aims to take the output of the previous phases and use it to create a detailed linguistic report. This report is generated using predefined templates. After the linguistic report is produced, a Large Language Model is utilised~\cite{brown2020} to further improve the report by providing more nuanced and descriptive information about the involved behaviours. The final report generated by this module includes essential observations and statements highlighting essential aspects of the observed behaviour. The methodology incorporates the Hierarchical Perception Mapping (HPM) and the Prompt Generator (PG) to evaluate the suspicion of disinformation in videos. Starting with the HPM, a recursive template that includes the linguistic label describing the evaluated phenomenon and its corresponding numerical value was defined. 

A prompt adjustment process based on contextual information is provided to guide the text generation, instructing the model to generate a comprehensive report using precise language. This approach has the advantage of producing output that avoids redundancies and highlights the most relevant aspects of the evaluation.

By leveraging the HPM and PG components, the methodology enhances the precision and relevance of generated prompts, enabling a more targeted and informative report on the suspicion of disinformation in videos. The hierarchical structure of the HPM allows us to capture nuanced details, ensuring that the prompts encompass the full range of pertinent information. The PG component optimises the output by emphasising key elements while maintaining coherence. Overall, this methodology harnesses the power of language models for prompt generation and facilitates the creation of comprehensive and insightful reports. Additionally, this methodology allows flexibility in adjusting the prompt according to specific needs. The expert can modify the prompt as desired, incorporating additional elements such as recommendations or warnings based on the results obtained. This customisation ensures that the generated reports align with the specific context of the problem being addressed. By providing the ability to tailor the prompts, the methodology empowers users to extract actionable insights and make informed decisions based on the evaluation of disinformation in videos. The model then presents a detailed, coherent, and clear explanation of the final result.

\section{Illustrative Example}
\label{sec:ilustrative_example}
In the model, 43 PMs are used of which 22 are first level (measures, from $1PM_1$ to $1PM_{22}$), 10 are second level (attributes, from $2PM_{23}$ to $2PM_{32}$), 5 are third level (dimensions, from $3PM_{33}$ to $3PM_{37}$), 5 are fourth level (behaviours, from $4PM_{38}$ to $4PM_{42}$) and 1 is fifth level (end result, $5PM_{43}$). 

\begin{figure}[ht!]
\centering
\includegraphics[scale=0.3]{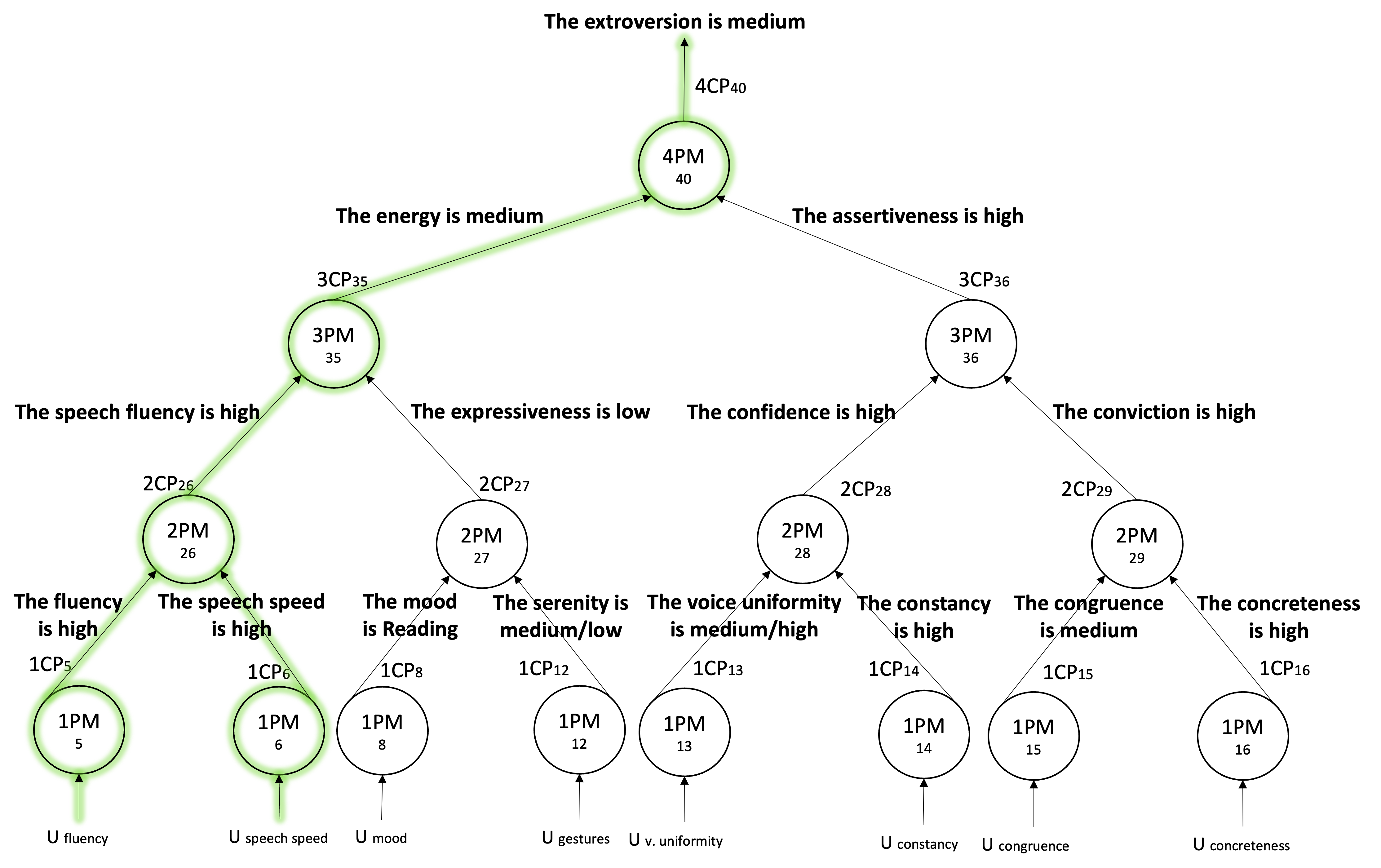}
\caption{GLMP for the extroversion.}
\label{fig:glmp_extroversion}
\end{figure}

An example of the use of GLMP for evaluating a 52-second video where the study subject is an adult person is described below \footnote{Link to the file: \url{https://github.com/smile-group/tiktok-disinformation/blob/main/videos_example/Video1.mp4}}. To present the evaluation process from the example video, the GLMP of the \textit{Extroversion} behaviour is shown in Fig. \ref{fig:glmp_extroversion} and the PMs highlighted in green are indicated in detail.

\begin{itemize}
\item \textbf{First-level Perception Mappings}
Each first-order PM accepts a single input $U$, reflecting a specific measure related to the study. Inputs can be normalised values between 0 and 10 or direct data source values. Based on these, each PM outputs a CP $y$, using a set of linguistic expressions $A$. Membership functions $g$ were generally defined for the labels \textit{Low}, \textit{Medium} and \textit{High} as (0, 0, 2.5, 5), (2.5, 5, 7.5) and (5, 7.5, 10, 10), respectively. 

Eight first-level PMs were defined for extroversion behaviour: speed, firmness, speech fluency, expressiveness, confidence, conviction, empathy and expression. The following describes the speech fluency and the speech speed attributes.

\textbf{Fluency} ($1PM_{5}$) linguistically describes the number of pauses in speech. It is a tuple $(U, y, g, T)$, where:
\begin{itemize}
    \item $U = 0$. The speech has no pauses. 
    \item $y$ is $1CP_{5} = (A,W)$ where $A=(high, medium, low)$ and $W = (1, 0, 0)$.
    \item $g$: the aggregation function is derived from a fuzzy partition comprising three membership functions: \textit{high} (trapezoidal membership function as (0, 0, 10, 20)), \textit{medium} (triangular membership function as (10, 20, 30)), and \textit{low} (trapezoidal membership function as (20, 30, $\infty, \infty$)). 
    \item $T$ is the template \textit{The fluency is high}.
\end{itemize}

\textbf{Speech speed} ($1PM_{6}$) linguistically describes the articulation in speech. It is a tuple $(U, y, g, T)$, where:
\begin{itemize}
    \item $U = 6.77$, which indicates the syllables per second. 
    \item $y$ is $1CP_{6} = (A,W)$ where $A=(low, medium, high)$ and $W = (0, 0, 1)$.
    \item $g$: the aggregation function is derived from a fuzzy partition comprising three membership functions: \textit{low} (trapezoidal membership function as (0, 0, 4, 5)), \textit{medium} (triangular membership function as (4, 5, 6)), and \textit{high} (trapezoidal membership function as (5, 6, $\infty, \infty$)). 
    \item $T$ is the template \textit{The speech speed is high}.
\end{itemize}

\item \textbf{Second-level Perception Mappings}
Second-order PMs receive multiple inputs from first-order PMs' outputs, thereby generating an output CP based on these inputs and predefined linguistic expressions. These expressions are as follows:

\begin{itemize}
    \item $A=(Bad, Normal, Good)$ for vocabulary, ideas, non-verbal communication, interaction tone and communication.
    \item $A=(Low, Medium, High)$ for speed, firmness, speech fluency, expressiveness, confidence, conviction, empathy and expression.
\end{itemize}

Membership degrees are consistent across all inputs. The aggregation function $g$ uses fuzzy rules for up to three inputs. For more than three, it uses a fuzzy weighted average. 

\textbf{Speech fluency} ($2PM_{26}$) describes the verbal agility. It is a tuple $(U, y, g, T)$, where:
\begin{itemize}
    \item $U = (1CP_{5},1CP_{6})$. 
    \item $y$ is $2CP_{26} = (A,W)$ where $A=(low, medium, high)$ and $W = (0, 0, 1)$.
    \item $g$: It is obtained using a set of nine fuzzy rules, for example:
    \begin{itemize}
        \renewcommand\labelitemi{}
        \item[] IF $1CP_{5}$ is \textit{low} and $1CP_{6}$ is \textit{low} THEN $2CP_{26}$ is \textit{low}.
        \item[] IF $1CP_{5}$ is \textit{low} and $1CP_{6}$ is \textit{medium} THEN $2CP_{26}$ is \textit{low}.
        \item[] ...
        \item[] IF $1CP_{5}$ is \textit{high} and $1CP_{6}$ is \textit{medium} THEN $2CP_{26}$ is \textit{high}.
        \item[] IF $1CP_{5}$ is \textit{high} and $1CP_{6}$ is \textit{high} THEN $2CP_{26}$ is \textit{high}.
    \end{itemize}
    \item $T$ is the template \textit{The speech fluency is high}.
\end{itemize}

\item \textbf{Third and fourth-level Perception Mappings} In both cases, each PM takes two CPs as inputs and outputs are defined with the same set of linguistic labels, $A = (Low, Medium, High)$. The corresponding membership degrees are (0, 0, 2.5, 5), (2.5, 5, 7.5) and (5, 7.5, 10,10), respectively, and the corresponding outputs are calculated using a set of fuzzy rules for the aggregation function $g$. The third-order PMs receive the inputs from the output CPs of the second-order PMs. 

\textbf{Energy} ($3PM_{35}$) describes the vigour of the person when communicating. It is a tuple $(U, y, g, T)$, where:
\begin{itemize}
    \item $U = (2CP_{26},2CP_{27})$. 
    \item $y$ is $3CP_{35} = (A,W)$ where $A=(low, medium, high)$ and $W = (0, 1, 0)$.
    \item $g$: It is obtained using a set of nine fuzzy rules, for example:
    \begin{itemize}
        \renewcommand\labelitemi{}
        \item[] IF $2CP_{26}$ is \textit{low} and $2CP_{27}$ is \textit{low} THEN $3CP_{35}$ is \textit{low}.
        \item[] IF $2CP_{26}$ is \textit{low} and $2CP_{27}$ is \textit{medium} THEN $3CP_{35}$ is \textit{low}.
        \item[] ...
        \item[] IF $2CP_{26}$ is \textit{high} and $2CP_{27}$ is \textit{medium} THEN $3CP_{35}$ is \textit{high}.
        \item[] IF $2CP_{26}$ is \textit{high} and $2CP_{27}$ is \textit{high} THEN $3CP_{35}$ is \textit{high}.
    \end{itemize}
    \item $T$ is the template \textit{The energy is medium}.
\end{itemize}

Still, depending on the specific behaviour being modelled, the fourth-order PMs can receive inputs from third-order PMs, second-order PMs, or a combination of both. 

\textbf{Extroversion} ($4PM_{40}$) describes the level of this behaviour. It is a tuple $(U, y, g, T)$, where:
\begin{itemize}
    \item $U = (3CP_{35},3CP_{36})$. 
    \item $y$ is $4CP_{40} = (A,W)$ where $A=(low, medium, high)$ and $W = (0, 1, 0)$.
    \item $g$: It is obtained using a set of nine fuzzy rules, for example:
    \begin{itemize}
        \renewcommand\labelitemi{}
        \item[] IF $3CP_{35}$ is \textit{low} and $3CP_{36}$ is \textit{low} THEN $4CP_{40}$ is \textit{low}.
        \item[] IF $3CP_{35}$ is \textit{low} and $3CP_{36}$ is \textit{medium} THEN $4CP_{40}$ is \textit{low}.
        \item[] ...
        \item[] IF $3CP_{35}$ is \textit{high} and $3CP_{36}$ is \textit{medium} THEN $4CP_{40}$ is \textit{high}.
        \item[] IF $3CP_{35}$ is \textit{high} and $3CP_{36}$ is \textit{high} THEN $4CP_{40}$ is \textit{high}.
    \end{itemize}
    \item $T$ is the template \textit{The extroversion is medium}.
\end{itemize}

\item \textbf{Fifth-level Perception Mapping}
In the model, the final stage involves a fifth-order perception mapping ($5PM_{43}$) that evaluates the suspicion of disinformation in a video. This mapping integrates the outputs of the five preceding fourth-order PMs, which represent behaviours. The aggregation function for this stage is the Choquet Integral (Eq. \ref{choquet_integral}), which considers the individual contribution of each behaviour and its interdependencies. This is achieved through a fuzzy measure that assigns weights to each behaviour and its combinations. In the example video, the speaker has openness at 4.03, conscientiousness at 5.47, extroversion at 5, agreeableness at 2.09 and neuroticism at 7.12. From these results, the degree of membership of each behaviour that describes a potential disinformer was calculated so that, in ascending order, the values to be added are agreeableness at 0, conscientiousness at 0.41, extroversion at 0.6, openness at 0.61 and neuroticism at 0.95. The application of the Choquet Integral yields an aggregated value of 0.65, which, following the membership functions, corresponds to a \textit{High} level of suspicion of disinformation.
Therefore, the output of this fifth-order PM, $5CP_{43}$, is a holistic and nuanced measure of the potential disinformation content in the video, considering both individual behaviours and their synergistic effects. 

\section{Experiments}
\label{sec:experiments}

TikTok, launched in 2016 by ByteDance and merged with Musical.ly in 2018, is a leading social media platform known for its short video content and recommendation algorithm. By 2022, it became the most downloaded app globally, with 672 million downloads. Its rapid growth, particularly among young users, has attracted scrutiny from media and fact-checking organisations due to the spread of misleading content, especially during the COVID-19 pandemic \cite{tiktok2022}.

The experiments were conducted to investigate the feasibility of the approach based on Deep Learning and Fuzzy Logic techniques to identify TikTok users suspected of generating disinformative content on the mentioned social network, which is the main research question. The latter was further analysed  into the following research questions:

\begin{itemize}
\item \textbf{RQ1}. Can disinformation users be identified in specific contexts using the proposed multimodal analysis of TikTok videos?
\item \textbf{RQ2}. Is it possible to scale up the proposed method to a topic that is widespread and has a high number of possible TikTok disinformation videos?
\end{itemize}

\subsection{Evaluation in specific contexts}
\label{subsec:evaluation_specific_contexts}

The dataset \cite{BhargavaDataset} from the study conducted by Bhargava et al. \cite{Bhargava2023} was used to validate in terms of identifying users who spread disinformation in specific contexts (\textbf{RQ1}). This study used a pre-registered survey experiment with 1,169 US participants to examine the impact of TikTok-style videos on six disinformation topics: aspartame causes cancer, the perceived danger of COVID-19, the Alec Baldwin shooting incident, natural immunity versus vaccinations, Ivermectin for COVID-19 treatment, and the distinction between left- and right-brained people. The sequence of videos presented was as follows: two disinformation videos, a correction video, and a debunking video, each presented for each topic, resulting in 24 videos, some of them  were omitted for the following reasons:

\begin{enumerate}
    \item Video 10: One disinformation video shows only the speaker's hands for 16 seconds out of 52.
    \item Video 11: This correction video includes numerous shaky camera shots of the sender, as he has recorded himself and does not maintain a fixed image.
    \item Video 12: Another correction video includes two distinct speeches.
    \item Video 16: The majority of the content in this debunking video is the video to be disproved.
    \item Video 23: It is invalid because it does not include a spoken presentation but rather a vocal performance.
\end{enumerate}

Table \ref{tab_validation} shows the results of the evaluation. Moreover, the results of the analysis processes have been published in an open access repository\footnote{Link to the file: \url{https://github.com/smile-group/tiktok-disinformation/blob/main/validation_results.csv}}.

\begin{table}[!htp]
    \caption{Evaluated videos for validation phase}
    \label{tab_validation}
    \centering
    \begin{tabular}{|c|>{\centering\arraybackslash}p{3cm}|c|c|>{\centering\arraybackslash}p{1.8cm}|}
     \hline
        \textbf{Video} & \textbf{Topic} & \textbf{Type} & \textbf{Suspect} & \textbf{Expected output} \\ \hline
        1 & \multirow{4}{3cm}{Aspartame causing cancer} & Disinformation & Medium & Yes \\
        2 & ~ & Disinformation & Medium & Yes \\
        3 & ~ & Correction & Low & Yes \\
        4 & ~ & Debunking & Low & Yes \\ \hline
        5 & \multirow{4}{3cm}{The perceived danger of COVID-19} & Disinformation & Low/Medium & No \\
        6 & ~ & Disinformation & Medium & Yes \\
        7 & ~ & Correction & Low/Medium & Yes \\
        8 & ~ & Debunking & Low & Yes \\ \hline
        9 & \multirow{4}{3cm}{Differentiating left vs. right-brained people} & Disinformation & Medium/High & Yes \\
        \(10^*\) & ~ & Disinformation & - & - \\
        \(11^*\) & ~ & Correction & - & - \\
        \(12^*\) & ~ & Debunking & - & - \\ \hline
        13 & \multirow{4}{3cm}{Alec Baldwin shooting incident} & Disinformation & High & Yes \\
        14 & ~ & Disinformation & Medium/High & Yes \\
        15 & ~ & Correction & Low & Yes\\
        \(16^*\) & ~ & Debunking & - & -\\ \hline
        17 & \multirow{4}{3cm}{Natural immunity versus vaccinations} & Disinformation & Medium/High & Yes \\
        18 & ~ & Disinformation & High & Yes \\
        19 & ~ & Correction & Low/Medium & Yes\\
        20 & ~ & Debunking & Low  & Yes\\ \hline
        21 & \multirow{4}{3cm}{Ivermectin for COVID-19 treatment} & Disinformation & High & Yes\\
        22 & ~ & Disinformation & Medium/High & Yes\\
        \(23^*\) & ~ & Correction & - &-\\
        24 & ~ & Debunking & Low  & Yes\\ \hline
    \end{tabular}
\end{table}

The results can be considered satisfactory as they confirmed that the correction videos contained low levels of disinformation, reaffirming their role in providing factual information. In addition, the debunking videos were also found to have low levels of disinformation, supporting their effectiveness in addressing and correcting false claims. Of the 19 videos evaluated, 18 achieved the expected results, demonstrating the robustness and reliability of the proposed approach in distinguishing between true and false information in TikTok-style speech videos.



\subsection{Evaluation in widespread contexts}
\label{subsec:evaluation_widespread_contexts}

Another experiment was conducted to validate the proposed approach in a widespread context (\textbf{RQ2}); in this case, the experiments analysed the video footage of individuals delivering speeches about the Invasion of Ukraine throughout 2022. For this purpose, a dataset generated by the Network Dynamics Lab at McGill University \cite{steel_benjamin_2023_7534952} was utilised, which compiled videos using specific keywords and hashtags related to the topic.

\subsubsection{Dataset Description}
The original dataset consists of 15,835 videos, 15 of which were removed as duplicates, leaving 15,820 videos spread across 10,176 users. This dataset extracted usernames and video IDs to identify videos that included close-up shots of the presenter. A video's eligibility for evaluation depended on two filters: its viability (as some videos might have been removed due to platform policies or account deletions) and the presence of spoken speech in the video. Of the 15,820 videos, 5,172 met these criteria and were considered eligible for evaluation.

\subsubsection{Data Processing} The processed dataset and the results, which include linguistic labels of behaviours and suspected disinformation, have been published in an open-access repository\footnote{Link to the file: \url{https://github.com/smile-group/tiktok-disinformation/blob/main/video_ids_results.csv}}. For videos that could not be assessed, their status is indicated as invalid. This category includes videos removed from TikTok while the user's account remains active and videos from permanently deleted accounts.

\subsubsection{Results}

The evaluation of 5,172 videos using the Big-5 behaviours resulted in the categorisation of suspected disinformation into five linguistic labels: \textit{Low}, \textit{Low/Medium}, \textit{Medium}, \textit{Medium/High}, and \textit{High}. The distribution of these labels is visualised in Fig. \ref{fig:distribution}.

\begin{itemize}
    \item Low suspicion: 831 videos (16.1\%) were categorized as having low suspicion of disinformation.
    \item Low/medium suspicion: This category included 1,127 videos (21.8\%), indicating a slightly higher suspicion level than the Low category.
    \item Medium suspicion: The largest group, with 1,318 videos (25.5\%), fell into the medium suspicion category, suggesting a moderate level of disinformation presence.
    \item Medium/high suspicion: A significant portion, 1,135 videos (22\%), was classified under medium/high suspicion, indicating a relatively high level of suspected disinformation.
    \item High suspicion: Finally, 761 videos (14.7\%) were identified as having a high suspicion of disinformation.
\end{itemize}

\begin{figure}[ht!]
\centering
\includegraphics[scale=0.45]{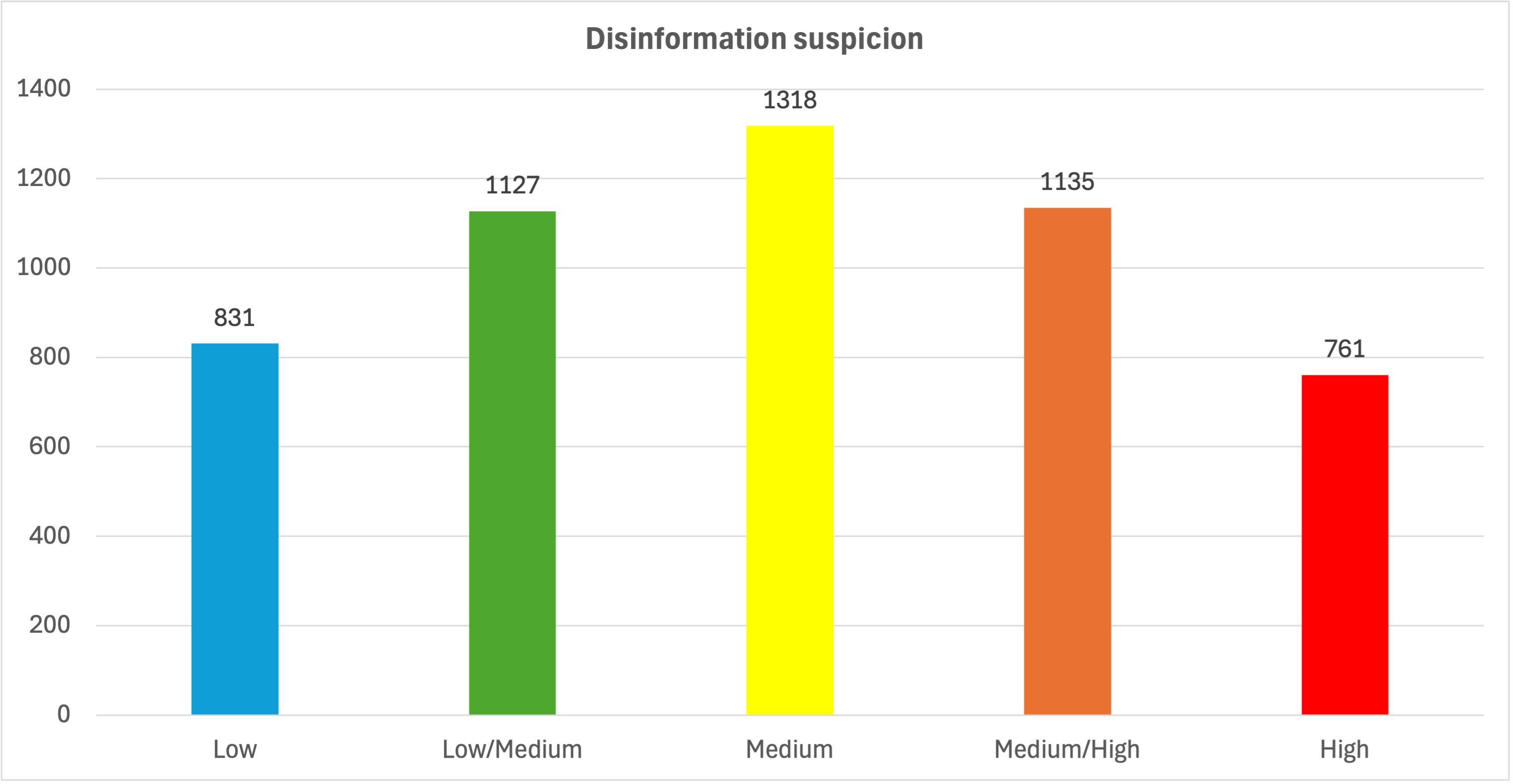}
\caption{Distribution of suspected disinformation from the evaluated videos}
\label{fig:distribution}
\end{figure}

Table~\ref{tab_results} shows a fragment of the results from the videos, where the labels that describe the behaviours and the level of suspicion of disinformation of the people evaluated are presented\footnote{Some videos have been removed from TikTok, but all can be viewed at: \url{https://github.com/smile-group/tiktok-disinformation/tree/main/videos_example}}. Results from the 5,172 videos have been made available in an open-access repository\footnote{Link to the file: \url{https://github.com/smile-group/tiktok-disinformation/blob/main/video_ids_results.csv}}.

Thus, a hierarchical structure report template has been developed that begins with the lowest level of detail (the degree of suspected disinformation) and progresses to the highest level (the representation of measures). Following the example video presented in Section \ref{sec:ilustrative_example} and the values obtained for the extroversion assessment, the report generated by the system is as follows. 

\begin{center}
\noindent\fbox{%
    \parbox{0.7\textwidth}{%

Extroversion is medium. Extroversion depends on energy and assertiveness. 
\begin{enumerate}
    \item Energy is medium. Energy depends on speech fluency and expressiveness.
     \begin{itemize}
         \item Speech fluency is high. Speech fluency depends on fluency and speech speed. 
         \begin{itemize}
             \item Fluency is high. 
             \item Speech speed is high. 
         \end{itemize}
         \item Expressiveness is low. Expressiveness depends on mood and serenity. 
         \begin{itemize}
             \item Mood is reading. 
             \item Serenity is medium. 
         \end{itemize}
     \end{itemize}
     \item Assertiveness is high. Assertiveness depends on confidence and conviction. 
     \begin{itemize}
         \item Confidence is high. Confidence depends on voice uniformity and constancy in expressing ideas. 
         \begin{itemize}
         \item Voice uniformity is medium/high. 
         \item Constancy in expressing ideas is high. 
         \end{itemize}
         \item Conviction is high. Conviction depends on congruence and concreteness.
         \begin{itemize}
         \item Congruence is medium/high. 
         \item Concreteness is high.
         \end{itemize}
     \end{itemize}
 \end{enumerate}    
    }%
}
\end{center}

It should be noted that each assessed behaviour follows the same structure and the outcome of suspected disinformation.

\begin{table}[!htp]
    \caption{Part of the results of the evaluation.}
    \label{tab_results}
    \centering
    \small
    \begin{tabular}{|p{0.5cm}|p{7 cm}|p{2.6cm}|}
     \hline
        \multicolumn{1}{|c}{Video}  & \multicolumn{1}{|c}{Behaviours} & 
            \multicolumn{1}{|c|}{Suspect}\\ \hline
            1 & High openness and extroversion, Low conscientiousness and agreeableness, Low/Medium neuroticism & Low/Medium  \\ 
            \hline
            2 & Medium openness, High conscientiousness and agreeableness, High extroversion, Low/Medium neuroticism & Medium\\
            \hline
            3 & Medium openness and neuroticism, Low/Medium conscientiousness, Low extroversion and agreeableness & Low/Medium \\
            \hline
            4 & High openness, conscientiousness and extroversion, Low agreeableness, Medium neuroticism & Low \\
            \hline
            5 & High openness, conscientiousness and extroversion, Low agreeableness, Medium neuroticism & Low \\
            \hline
            6 & Medium openness and extroversion, Medium/High conscientiousness, Low agreeableness, High neuroticism & High\\
            \hline
            7 & High openness, conscientiousness and extroversion, Low agreeableness, Medium neuroticism & Low \\
            \hline
            8 & Medium/High openness, Low conscientiousness, extroversion and agreeableness, Medium neuroticism & Medium\\
            \hline
            9 & High openness and neuroticism, Medium/High conscientiousness, Low/Medium extroversion, Low agreeableness & High \\
            \hline
            10 & Medium openness and neuroticism, Low conscientiousness, Low/Medium extroversion, High agreeableness & Low\\
        \hline
    \end{tabular}
\end{table}

The level of detail of the explanations generated with the template is improved using LLM, specifically OpenAI's Generative PreTrained model (GPT-4). Tables~\ref{tab_report_low},~\ref{tab_report_medium} and~\ref{tab_report_high}  show the provided report for this example for a low, medium and high suspicion of disinformation, respectively.

\begin{table}[!htp]
\caption{Example of results of low suspected disinformation.}
\centering
\footnotesize
\begin{tabular}{|c| >{\raggedright\arraybackslash}p{12cm}|}
\hline
Video & \multicolumn{1}{|c|}{Natural Language Report} \\
\hline
4 & The person who sent the message has a \textit{low} probability of spreading disinformation, as they are \textit{not very} \textbf{open}, \textit{not very} \textbf{conscious}, \textit{not very} \textbf{extroverted}, \textit{not very} \textbf{agreeable}, and \textbf{emotionally stable} \textit{on average}. The person is \textit{not very} \textbf{open} because the \textbf{quality of the text} is \textit{low} and the \textbf{nonverbal language} is \textit{normal}. The person is \textit{not very} \textbf{conscious} because the \textbf{conciseness} and \textbf{clarity} of the message are \textit{low/medium}. The person is \textit{not very} \textbf{extroverted} as they have \textit{low} \textbf{energy} and \textbf{expressiveness}, although they are \textit{high} in \textbf{assertiveness}. The person is \textit{not very} \textbf{agreeable} as they have \textit{low} \textbf{empathy} and the \textbf{tone of interaction} is \textit{poor}. \textit{On average}, the person has a \textit{medium} level of \textbf{neuroticism}, and their \textbf{emotional stability} is \textit{medium/high}. Based on the presented results, the level of suspicion regarding the person in the video spreading disinformation is \textit{low}.\\
\hline
\end{tabular}
\label{tab_report_low}
\end{table}

\begin{table}[!h]
\caption{Example of results of medium suspected disinformation.}
\centering
\footnotesize
\begin{tabular}{|c| >{\raggedright\arraybackslash}p{12cm}|}
\hline
Video & \multicolumn{1}{|c|}{Natural Language Report} \\
\hline
2 & The person who sent the message has a \textit{medium} probability of spreading disinformation, as they are \textit{moderately} \textbf{open}, \textit{highly} \textbf{conscious}, \textit{highly} \textbf{extroverted}, \textit{highly} \textbf{agreeable}, and \textbf{emotionally stable} \textit{on average}. The person is \textit{moderately} \textbf{open} because the \textbf{quality of the text} is \textit{medium}, while the \textbf{nonverbal language} is \textit{normal/good} considering blinking, gaze, facial movements, and gestures. The person is \textit{highly} \textbf{conscious} as the \textbf{conciseness} and \textbf{clarity} of the message are \textit{high}. The person demonstrates \textit{high} \textbf{energy} in their \textbf{expressiveness} and \textit{high} \textbf{fluency} in their speech, although the \textbf{use of examples} is \textit{medium}. The person is \textit{highly} \textbf{assertive} due to their \textit{high} \textbf{confidence} and \textbf{conviction}. They exhibit \textit{high} \textbf{empathy}, although the \textbf{smile} is scarce, and their \textbf{gaze} is \textit{centred/normal}. The \textbf{tone of interaction} is \textit{good}, with \textit{neutral} \textbf{polarity} and \textit{high} \textbf{voice uniformity}. \textit{On average}, the person has a \textit{low/medium} level of \textbf{neuroticism}, and their \textbf{emotional stability} is \textit{medium}. Based on the presented results, the level of suspicion regarding the person in the video spreading disinformation is \textit{medium}.\\
\hline
\end{tabular}
\label{tab_report_medium}
\end{table}

\begin{table}[ht!]
\caption{Example of results of high suspected disinformation.}
\centering
\footnotesize
\begin{tabular}{|c| >{\raggedright\arraybackslash}p{12cm}|}
\hline
Video & \multicolumn{1}{|c|}{Natural Language Report} \\
\hline
6 & The person who sent the message has a \textit{high} probability of spreading disinformation, as they are \textit{moderately} \textbf{open}, \textit{highly} \textbf{conscious}, \textit{medium} \textbf{extroverted}, \textit{low} \textbf{agreeable}, and \textbf{emotionally stable} \textit{on average}. The person is \textit{medium} \textbf{open} because the \textbf{quality of the text} is \textit{medium}, while the \textbf{nonverbal language} is \textit{good}, considering the tone of voice used for reading, without taking into account blinking, gaze, posture changes, and gestures. The person is \textit{highly} \textbf{conscious} as the \textbf{conciseness} and \textbf{clarity} of the message are \textit{high}. The person demonstrates \textit{medium} \textbf{energy} in their \textbf{fluency} of speech, although the \textbf{expressiveness} is \textit{low}. They exhibit \textit{high} \textbf{confidence} in their \textbf{assertiveness}, while their \textbf{conviction} is \textit{high}. The person displays \textit{low} \textbf{empathy}, with a scarce \textbf{smile}, and a \textbf{normal/diffuse} gaze. The \textbf{tone of interaction} is \textit{poor}, with \textit{negative} \textbf{polarity} and \textit{medium} \textbf{voice uniformity}. Regarding \textbf{neuroticism}, the person has a \textit{high} level, with \textit{medium} control of stress and \textit{low/medium} emotional stability.\\
\hline
\end{tabular}
\label{tab_report_high}
\end{table}
\end{itemize}

\subsection{Discussion}

The findings of the initial assessment (Section~\ref{subsec:evaluation_specific_contexts}), based on a dataset 
that contains several disinformation subjects, including health-related misinformation and social incidents, demonstrate that the proposed model, through the utilisation of multimodal features extracted from TikTok videos, can accurately identify individuals engaged in disseminating disinformation in context-limited environments based on their behavioural patterns. (\textbf{RQ1}).

The second evaluation, described in Section~\ref{subsec:evaluation_widespread_contexts}, focused on the scalability of the proposed model (\textbf{RQ2}), examining its performance on a more extensive dataset encompassing the Invasion of Ukraine, a subject of considerable contemporary relevance and a high volume of potential disinformation. 

The results of this experiment illustrate the efficacy of the proposed hybrid methodology in discerning potential disinformative content on TikTok videos. In particular, the methodology classified 16.1\% of videos as low-risk, 21.8\% as low-to-medium-risk, 25.5\% as medium-risk, 22\% as medium-to-high-risk, and 14.7\% as high-risk. This distribution highlights the model’s capacity to differentiate between varying degrees of disinformation, thereby providing a detailed insight into the quality of content on the platform.

The multimodal feature analyser's performance demonstrated robust capabilities in processing and evaluating a wide range of data formats. Text analysis effectively identified features such as vocabulary redundancy and text coherence, while audio analysis captured speech fluency and emotional tone. Video analysis corroborated these findings by examining non-verbal cues like facial expressions and body language.
These multimodal insights contribute to a more accurate assessment of disinformation. The method’s ability to translate these various forms of data into coherent and actionable results highlights the advantage of a comprehensive, multimodal approach in tackling the complexities of disinformation detection.

\subsection{Future Challenges}
Several challenges emerge from the results:
\begin{itemize}
    \item Metric dependency: The system’s reliance on predefined metrics—such as position, duration, and language—affected the depth of analysis. While valuable, these metrics have limitations in that they can result in the loss of contextual knowledge, affecting detection accuracy. Future improvements could include refining or expanding these metrics to capture more disinformation indicators.
    \item Translation issues: The necessity for double translation in languages other than English, such as Russian, presented significant challenges. This underscores the importance of improving translation models to ensure the accurate processing and analysis of non-English content, thereby enhancing the system’s overall effectiveness
    \item Deepfake detection: The current methodology does not address the issue of deepfake detection, which restricts its application to textual and auditory analysis. Developing techniques for detecting deepfakes and image forgeries is a significant area for future research. The integration of visual forgery detection techniques could provide a more comprehensive approach to the detection of disinformation.
    \item Complexity of multimodal datasets: The processing and integration of multimodal data present several challenges, particularly concerning data synchronisation and aggregation. In the format used by TikTok, resources such as subtitles are employed to highlight key points in a video. However, when these assets are distributed across most of the video and placed in a position that obscures the elements to be analysed, such as the face, eyes and mouth, they become a significant challenge for analysis. This is because non-verbal language and facial expressions may not be evaluated appropriately. 
\end{itemize}

The identified limitations highlight areas for improvement. Addressing these issues, such as enhancing translation accuracy, incorporating deepfake detection, and refining metric definitions, will strengthen the methodology and expand its utility. Integrating expert knowledge with advanced computational techniques offers a promising foundation; nevertheless, continuous refinement and adaptation are necessary to keep pace with evolving disinformation tactics.

\section{Conclusions}
\label{sec:concl}
This research introduces a novel methodology for analysing disinformation patterns within videos, emphasising uncovering common behaviours typically seen in disinformation content. The primary contribution is a hybrid intelligence framework designed explicitly for TikTok, which merges Deep Learning's computational prowess and Fuzzy Logic's interpretability to identify multimodal disinformation effectively. This approach integrates various disciplines, including computer vision, audio and text analysis, psychology, and Fuzzy Logic, offering a comprehensive and accurate understanding of disinformation behaviours. Consequently, this work paves the way for a deeper comprehension of disinformation trends and significantly boosts the capability to detect common disinformation traits in videos.

The application of the framework to a dataset of TikTok videos yielded promising results in disinformation detection, highlighting its potential as a valuable asset in restraining the spread of false information. In addition to successful detection, the system generates high-quality, comprehensive, and well-structured reports, providing a detailed view of the disinformation behaviours in question. Furthermore, a standout feature of the system is its inherent explainability and traceability. Each step of the detection process is explained, offering transparency and allowing users to understand the rationale behind the identified disinformation content. Therefore, the system provides a potent combination of precise detection and comprehensive, traceable explanations, marking it as a significant tool in understanding and combating disinformation on multimedia platforms. Another contribution of this work is a dataset specifically labelled and designed to include the results of human behavioural assessment, focusing on identifying signals that may suggest the presence of disinformation.

Future research will enhance the hybrid intelligence strategy by incorporating additional cognitive computing elements, like sentiment analysis and NLP, into the model. This augmentation could increase the model's accuracy and sensitivity, allowing it to detect and interpret more complex forms of disinformation, often involving factual inaccuracies and emotional manipulation. In addition, it is planned to extend testing of the model to a broader range of multimedia platforms, including YouTube, Facebook, Instagram, and X. By understanding each platform's unique platform in disinformation detection; the model can be adjusted to improve its effectiveness and versatility across a wide array of digital media platforms.

\section*{Availability of data and materials}
\small
\noindent The results of the experiments are available in the following GitHub repository: \\ \url{https://github.com/smile-group/tiktok-disinformation}

\section*{Acknowledgements}
\small
\noindent The Spanish Government has partially supported this work under the grant SAFER: PID2019-104735RB-C42 (ERA/ERDF, EU), and project XAI-DISINFODEMICS (PLEC2021-007681) funded by MCIN/AEI /10.13039/501100011033 and by the European Union NextGenerationEU/ PRTR.

\bibliographystyle{unsrt}  
\bibliography{references}  

\begin{thebibliography}{10}

\bibitem{Muhammed2022}
Sadiq Muhammed~T and Saji~K Mathew.
\newblock The disaster of misinformation: a review of research in social media.
\newblock {\em International journal of data science and analytics}, 13(4):271--285, 2022.

\bibitem{delVicario2016}
Michela Del~Vicario, Alessandro Bessi, Fabiana Zollo, Fabio Petroni, Antonio Scala, Guido Caldarelli, H~Eugene Stanley, and Walter Quattrociocchi.
\newblock The spreading of misinformation online.
\newblock {\em Proceedings of the national academy of Sciences}, 113(3):554--559, 2016.

\bibitem{Su2016}
Lin Su and Martin Levine.
\newblock {Does “lie to me” lie to you? An evaluation of facial clues to high-stakes deception}.
\newblock {\em Computer Vision and Image Understanding}, 147:52--68, jun 2016.

\bibitem{Nguyen2022}
Thanh~Thi Nguyen, Quoc Viet~Hung Nguyen, Dung~Tien Nguyen, Duc~Thanh Nguyen, Thien Huynh-The, Saeid Nahavandi, Thanh~Tam Nguyen, Quoc-Viet Pham, and Cuong~M Nguyen.
\newblock {Deep learning for deepfakes creation and detection: A survey}.
\newblock {\em Computer Vision and Image Understanding}, 223:103525, 2022.

\bibitem{choi2022}
Hyewon Choi and Youngjoong Ko.
\newblock Effective fake news video detection using domain knowledge and multimodal data fusion on youtube.
\newblock {\em Pattern Recognition Letters}, 154:44--52, 2022.

\bibitem{Bonomi2021}
Mattia Bonomi, Cecilia Pasquini, and Giulia Boato.
\newblock Dynamic texture analysis for detecting fake faces in video sequences.
\newblock {\em Journal of Visual Communication and Image Representation}, 79:103239, 2021.

\bibitem{Shu2017}
Kai Shu, Amy Sliva, Suhang Wang, Jiliang Tang, and Huan Liu.
\newblock Fake news detection on social media: A data mining perspective.
\newblock {\em SIGKDD Explor. Newsl.}, 19(1):22–36, sep 2017.

\bibitem{Guerrero-Sosa2023}
Jared D.~T. Guerrero-Sosa, Francisco~P. Romero, Andres Montoro-Montarroso, Victor~H. Menendez, Jesus Serrano-Guerrero, and Jose~A. Olivas.
\newblock A fuzzy approach to detecting suspected disinformation in videos.
\newblock In Henrik~Legind Larsen, Maria~J Martin-Bautista, M~Dolores Ruiz, Troels Andreasen, Gloria Bordogna, and Guy~De Tré, editors, {\em Flexible Query Answering Systems}, pages 145--158, 2023.

\bibitem{Comito2023}
Carmela Comito, Luciano Caroprese, and Ester Zumpano.
\newblock Multimodal fake news detection on social media: a survey of deep learning techniques.
\newblock {\em Social Network Analysis and Mining}, 13:101, 2023.

\bibitem{Tufchi2023}
Shivani Tufchi, Ashima Yadav, and Tanveer Ahmed.
\newblock A comprehensive survey of multimodal fake news detection techniques: advances, challenges, and opportunities.
\newblock {\em International Journal of Multimedia Information Retrieval}, 12:28, 2023.

\bibitem{Wilson2023}
Anna Wilson, Seb Wilkes, Yayoi Teramoto, and Scott Hale.
\newblock Multimodal analysis of disinformation and misinformation.
\newblock {\em Royal Society Open Science}, 10(12):230964, 2023.

\bibitem{MontoroMontarroso2023}
Andrés Montoro-Montarroso, Javier Cantón-Correa, Paolo Rosso, Berta Chulvi, Ángel Panizo-Lledot, Javier Huertas-Tato, Blanca Calvo-Figueras, M.~José Rementeria, and Juan Gómez-Romero.
\newblock Fighting disinformation with artificial intelligence: fundamentals, advances and challenges.
\newblock {\em El Profesional de la información}, 6 2023.

\bibitem{Schuster2020}
Tal Schuster, Roei Schuster, Darsh~J. Shah, and Regina Barzilay.
\newblock The limitations of stylometry for detecting machine-generated fake news.
\newblock {\em Computational Linguistics}, 46:499--510, 6 2020.

\bibitem{PENG2023120501}
Zifan Peng, Mingchen Li, Yue Wang, and George~T.S. Ho.
\newblock Combating the covid-19 infodemic using prompt-based curriculum learning.
\newblock {\em Expert Systems with Applications}, 229:120501, 2023.

\bibitem{Rastogi2022}
Shubhangi Rastogi and Divya Bansal.
\newblock Disinformation detection on social media: An integrated approach.
\newblock {\em Multimedia Tools and Applications}, 81:40675--40707, 2022.

\bibitem{Hangloo2022}
Sakshini Hangloo and Bhavna Arora.
\newblock Combating multimodal fake news on social media: methods, datasets, and future perspective.
\newblock {\em Multimedia Systems}, 28:2391--2422, 12 2022.

\bibitem{singh2021}
Vivek~K. Singh, Isha Ghosh, and Darshan Sonagara.
\newblock Detecting fake news stories via multimodal analysis.
\newblock {\em Journal of the Association for Information Science and Technology}, 72(1):3--17, 2021.

\bibitem{CHAI2024121588}
Yidong Chai, Yi~Liu, Weifeng Li, Bin Zhu, Hongyan Liu, and Yuanchun Jiang.
\newblock An interpretable wide and deep model for online disinformation detection.
\newblock {\em Expert Systems with Applications}, 237:121588, 2024.

\bibitem{Sengan2023}
Sudhakar Sengan, Subramaniyaswamy Vairavasundaram, Logesh Ravi, Ahmad Qasim~Mohammad AlHamad, Hamzah~Ali Alkhazaleh, and Meshal Alharbi.
\newblock Fake news detection using stance extracted multimodal fusion-based hybrid neural network.
\newblock {\em IEEE Transactions on Computational Social Systems}, pages 1--12, 2023.

\bibitem{Jing2023}
Jing Jing, Hongchen Wu, Jie Sun, Xiaochang Fang, and Huaxiang Zhang.
\newblock Multimodal fake news detection via progressive fusion networks.
\newblock {\em Information Processing \& Management}, 60:103120, 1 2023.

\bibitem{Ghorbanpour2023}
Faeze Ghorbanpour, Maryam Ramezani, Mohammad~Amin Fazli, and Hamid~R. Rabiee.
\newblock Fnr: a similarity and transformer-based approach to detect multi-modal fake news in social media.
\newblock {\em Social Network Analysis and Mining}, 13:56, 3 2023.

\bibitem{Guo2023}
Ying Guo.
\newblock A mutual attention based multimodal fusion for fake news detection on social network.
\newblock {\em Applied Intelligence}, 53:15311--15320, 6 2023.

\bibitem{Yadav2023}
Ashima Yadav, Shivani Gaba, Haneef Khan, Ishan Budhiraja, Akansha Singh, and Krishna~Kant Singh.
\newblock Etma: Efficient transformer-based multilevel attention framework for multimodal fake news detection.
\newblock {\em IEEE Transactions on Computational Social Systems}, pages 1--13, 2023.

\bibitem{Hua2023}
Jiaheng Hua, Xiaodong Cui, Xianghua Li, Keke Tang, and Peican Zhu.
\newblock Multimodal fake news detection through data augmentation-based contrastive learning.
\newblock {\em Applied Soft Computing}, 136:110125, 3 2023.

\bibitem{Meel2023}
Priyanka Meel and Dinesh~Kumar Vishwakarma.
\newblock Multi-modal fusion using fine-tuned self-attention and transfer learning for veracity analysis of web information.
\newblock {\em Expert Systems with Applications}, 229:120537, 11 2023.

\bibitem{Xiong2023}
Shufeng Xiong, Guipei Zhang, Vishwash Batra, Lei Xi, Lei Shi, and Liangliang Liu.
\newblock Trimoon: Two-round inconsistency-based multi-modal fusion network for fake news detection.
\newblock {\em Information Fusion}, 93:150--158, 5 2023.

\bibitem{Singh2023}
Prabhav Singh, Ridam Srivastava, K.P.S. Rana, and Vineet Kumar.
\newblock Semi-fnd: Stacked ensemble based multimodal inferencing framework for faster fake news detection.
\newblock {\em Expert Systems with Applications}, 215:119302, 4 2023.

\bibitem{Wu2023}
Lianwei Wu, Pusheng Liu, Yongqiang Zhao, Peng Wang, and Yangning Zhang.
\newblock Human cognition-based consistency inference networks for multi-modal fake news detection.
\newblock {\em IEEE Transactions on Knowledge and Data Engineering}, 14:1--14, 2023.

\bibitem{Giachanou2020}
Anastasia Giachanou, Guobiao Zhang, and Paolo Rosso.
\newblock Multimodal fake news detection with textual, visual and semantic information.
\newblock In Petr Sojka, Ivan Kope{\v{c}}ek, Karel Pala, and Ale{\v{s}} Hor{\'a}k, editors, {\em Text, Speech, and Dialogue}, pages 30--38, Cham, 2020. Springer International Publishing.

\bibitem{Li2022}
Shuo Li, Tao Yao, Saifei Li, and Lianshan Yan.
\newblock Semantic‐enhanced multimodal fusion network for fake news detection.
\newblock {\em International Journal of Intelligent Systems}, 37:12235--12251, 12 2022.

\bibitem{Zhang2022}
Guobiao Zhang, Anastasia Giachanou, and Paolo Rosso.
\newblock Scenefnd: Multimodal fake news detection by modelling scene context information.
\newblock {\em Journal of Information Science}, pages 355--367, 4 2022.

\bibitem{Wang2023}
Bin Wang, Yong Feng, Xian cai Xiong, Yong heng Wang, and Bao hua Qiang.
\newblock Multi-modal transformer using two-level visual features for fake news detection.
\newblock {\em Applied Intelligence}, 53:10429--10443, 5 2023.

\bibitem{Silva2023}
Marcos Paulo~Silva Gôlo, Mariana~Caravanti de~Souza, Rafael~Geraldeli Rossi, Solange~Oliveira Rezende, Bruno~Magalhães Nogueira, and Ricardo~Marcondes Marcacini.
\newblock One-class learning for fake news detection through multimodal variational autoencoders.
\newblock {\em Engineering Applications of Artificial Intelligence}, 122:106088, 6 2023.

\bibitem{Hameleers2020}
Michael Hameleers, Thomas~E. Powell, Toni G.L.A. Van~Der Meer, and Lieke Bos.
\newblock A picture paints a thousand lies? the effects and mechanisms of multimodal disinformation and rebuttals disseminated via social media.
\newblock {\em Political Communication}, 37(2):281--301, 2020.

\bibitem{Chen2023}
Jinyin Chen, Chengyu Jia, Haibin Zheng, Ruoxi Chen, and Chenbo Fu.
\newblock Is multi-modal necessarily better? robustness evaluation of multi-modal fake news detection.
\newblock {\em IEEE Transactions on Network Science and Engineering}, pages 1--15, 2023.

\bibitem{Han2024}
Linfeng Han, Xiaoming Zhang, Ziyi Zhou, and Yun Liu.
\newblock A multifaceted reasoning network for explainable fake news detection.
\newblock {\em Information Processing \& Management}, 61(6):103822, 2024.

\bibitem{Peng2024}
Liwen Peng, Songlei Jian, Zhigang Kan, Linbo Qiao, and Dongsheng Li.
\newblock Not all fake news is semantically similar: Contextual semantic representation learning for multimodal fake news detection.
\newblock {\em Information Processing \& Management}, 61(1):103564, 2024.

\bibitem{Huda2024}
Noor ul~Huda, Ali Javed, Kholoud Maswadi, Ali Alhazmi, and Rehan Ashraf.
\newblock Fake-checker: A fusion of texture features and deep learning for deepfakes detection.
\newblock {\em Multimedia Tools and Applications}, 83:49013--49037, 2024.

\bibitem{zadeh2002granular}
Lotfi~A Zadeh.
\newblock Granular computing as a basis for a computational theory of perceptions.
\newblock In {\em 2002 IEEE World Congress on Computational Intelligence. 2002 IEEE International Conference on Fuzzy Systems. FUZZ-IEEE'02. Proceedings (Cat. No. 02CH37291)}, volume~1, pages 564--565. IEEE, 2002.

\bibitem{TRIVINO201322}
Gracian Trivino and Michio Sugeno.
\newblock Towards linguistic descriptions of phenomena.
\newblock {\em International Journal of Approximate Reasoning}, 54(1):22--34, 2013.

\bibitem{CONDECLEMENTE201746}
Patricia Conde-Clemente, Jose~M. Alonso, Éldman O.~Nunes, Angel Sanchez, and Gracian Trivino.
\newblock New types of computational perceptions: Linguistic descriptions in deforestation analysis.
\newblock {\em Expert Systems with Applications}, 85:46--60, 2017.

\bibitem{de2022natural}
Andrea de~Anda-Trasvi{\~n}a, Alejandra Nieto-Garibay, and Joaqu{\'\i}n Guti{\'e}rrez.
\newblock Natural language report of the composting process status using linguistic perception.
\newblock {\em Applied Soft Computing}, 127:109357, 2022.

\bibitem{novikova2019five}
Irina~A Novikova and Alexandra~A Vorobyeva.
\newblock The five-factor model: Contemporary personality theory.
\newblock {\em Cross-Cultural Psychology: Contemporary Themes and Perspectives}, pages 685--706, 2019.

\bibitem{Radford2022b}
Alec Radford, Jong~Wook Kim, Christine McLeavey, Pamela Mishkin, Tao Xu, Greg Brockman, and Ilya Sutskever.
\newblock {Introducing Whisper}, 2022.

\bibitem{Chen2018}
Lei Chen, Klaus Zechner, Su-Youn Yoon, Keelan Evanini, Xinhao Wang, Anastassia Loukina, Jidong Tao, Lawrence Davis, Chong~Min Lee, Min Ma, Robert Mundkowsky, Chi Lu, Chee~Wee Leong, and Binod Gyawali.
\newblock Automated scoring of nonnative speech using the speechratersm v. 5.0 engine.
\newblock {\em ETS Research Report Series}, 2018(1):1--31, 2018.

\bibitem{mikolov2017advances}
Tomas Mikolov, Edouard Grave, Piotr Bojanowski, Christian Puhrsch, and Armand Joulin.
\newblock Advances in pre-training distributed word representations, 2017.

\bibitem{Pennington2014}
Jeffrey Pennington, Richard Socher, and Christopher~D. Manning.
\newblock {GloVe: Global Vectors for Word Representation}.
\newblock In {\em Proceedings of the 2014 Conference on Empirical Methods in Natural Language Processing (EMNLP)}, pages 1532--1543, Doha, 2014. Association for Computational Linguistics.

\bibitem{Martinez-Castano2020}
Rodrigo Martínez-Castaño, Juan~C. Pichel, and David~E. Losada.
\newblock A big data platform for real time analysis of signs of depression in social media.
\newblock {\em International Journal of Environmental Research and Public Health}, 17(13), 2020.

\bibitem{Ermakova2023}
Tatiana Ermakova, Benjamin Fabian, Elena Golimblevskaia, and Max Henke.
\newblock A comparison of commercial sentiment analysis services.
\newblock {\em SN Computer Science}, 4:477, 2023.

\bibitem{GIANNAKAKIS201789}
G~Giannakakis, M~Pediaditis, D~Manousos, E~Kazantzaki, F~Chiarugi, P~G Simos, K~Marias, and M~Tsiknakis.
\newblock {Stress and anxiety detection using facial cues from videos}.
\newblock {\em Biomedical Signal Processing and Control}, 31:89--101, 2017.

\bibitem{6909637}
Vahid Kazemi and Josephine Sullivan.
\newblock One millisecond face alignment with an ensemble of regression trees.
\newblock In {\em 2014 IEEE Conference on Computer Vision and Pattern Recognition}, pages 1867--1874, 2014.

\bibitem{6909616}
Yaniv Taigman, Ming Yang, Marc'Aurelio Ranzato, and Lior Wolf.
\newblock Deepface: Closing the gap to human-level performance in face verification.
\newblock In {\em 2014 IEEE Conference on Computer Vision and Pattern Recognition}, pages 1701--1708, 2014.

\bibitem{Goldberg1992}
L.R. Goldberg.
\newblock {The development of markers for the Big-Five factor structure}.
\newblock {\em Psychological Assessment}, 4(1):26--42, 1992.

\bibitem{PYKL2022117952}
P.Y.K.L. Srinivas, Amitava Das, and Viswanath Pulabaigari.
\newblock Fake spreader is a narcissist; real spreader is machiavellian prediction of fake news diffusion using psycho-sociological facets.
\newblock {\em Expert Systems with Applications}, 207:117952, 2022.

\bibitem{Beliakov2016}
Gleb Beliakov, Humberto Bustince~Sola, and Tomasa Calvo~S{\'a}nchez.
\newblock {\em Fuzzy Integrals}, chapter~4, pages 145--181.
\newblock Springer International Publishing, Cham, 2016.

\bibitem{brown2020}
Tom~B. Brown, Benjamin Mann, Nick Ryder, Melanie Subbiah, and al.
\newblock Language models are few-shot learners, 2020.

\bibitem{tiktok2022}
Sim{\'o}n Pe{\~n}a-Fern{\'a}ndez, Ainara Larrondo-Ureta, and Jordi Morales-i Gras.
\newblock Current affairs on tiktok. virality and entertainment for digital natives.
\newblock {\em Profesional de la Informaci{\'o}n}, 31(1), 2022.

\bibitem{BhargavaDataset}
Puneet Bhargava, Katie MacDonald, Christie Newton, Hause Lin, and Gordon Pennycook.
\newblock How effective are tiktok misinformation debunking videos?, October 2022.

\bibitem{Bhargava2023}
Puneet Bhargava, Katie MacDonald, Christie Newton, Hause Lin, and Gordon Pennycook.
\newblock How effective are tiktok misinformation debunking videos?
\newblock {\em Harvard Kennedy School Misinformation Review}, 3 2023.

\bibitem{steel_benjamin_2023_7534952}
Benjamin Steel, Sara Parker, and Derek Ruths.
\newblock The invasion of ukraine viewed through tiktok: A dataset, January 2023.

\end{thebibliography}

\end{document}